# Dynamically configured physics-informed neural network in topology optimization applications

Jichao Yin[a,b*]; Ziming Wen[a,b]; Shuhao Li[a]; Yaya Zhang[a]; Hu Wang[a,b†]

[a] *State Key Laboratory of Advanced Design and Manufacturing for Vehicle, Hunan University, Changsha, 410082, People's Republic of China*

[b] *Beijing Institute of Technology Shenzhen Automotive Research Institute, Shenzhen, 518000, People's Republic of China*

## Abstract

Integration of machine learning (ML) into the topology optimization (TO) framework is attracting increasing attention, but data acquisition in data-driven models is prohibitive. Compared with popular ML methods, the physics-informed neural network (PINN) can avoid generating enormous amounts of data when solving forward problems and additionally provide better inference. To this end, a dynamically configured PINN-based topology optimization (DCPINN-TO) method is proposed. The DCPINN is composed of two subnetworks, namely the backbone neural network (NN) and the coefficient NN, where the coefficient NN has fewer trainable parameters. The designed architecture aims to dynamically configure trainable parameters; that is, an inexpensive NN is used to replace an expensive one at certain optimization cycles. Furthermore, an active sampling strategy is proposed to selectively sample collocations depending on the pseudo-densities at each optimization cycle. In this manner, the number of collocations will decrease with the optimization process but will hardly affect it. The Gaussian integral is used to calculate the strain energy of elements, which yields a byproduct of decoupling the mapping of the material at the collocations. Several examples with different resolutions validate the feasibility of the DCPINN-TO method, and multiload and multiconstraint problems are employed to illustrate its generalization. In addition, compared to finite element analysis-based TO (FEA-TO), the accuracy of the displacement prediction and optimization results indicate that the DCPINN-TO method is effective and efficient.

[*] First author. E-mail address: jcyin@hnu.edu.cn (J. Yin)

[†] Corresponding author. E-mail address: wanghu@hnu.edu.cn (H. Wang)

# 1. Introduction

Topology optimization (TO) [1] can generate elegant designs by managing the spatial distribution of materials, with the advantage that the topology of the structure can be changed and the potential of materials can be explored. TO has become a well-established discipline with successful applications in aerospace [2], automotive [3], and other fields. After decades of rapid development, popular methods such as the solid isotropic material with penalization (SIMP) method [4], the evolutionary method [5], the level sets method [6], and integer programming-based methods [7,8] have been developed thus far. The element-wise/node-wise paradigm allows TO to have a larger design space than traditional structural optimization methods. However, the larger design space also implies more expensive computational costs, especially with the nested structural response analysis inherent in structural optimization methods [9]. For this reason, various methods have been proposed in an attempt to improve the computational efficiency of TO, such as the multi-grid conjugate gradient method [10], the reduced order models [11,12], and parallel techniques [13], etc. Furthermore, recent advances in the integration of computational mechanics and machine learning (ML) provide a new perspective [14], and research on integrating ML into TO frameworks has attracted increasing attention.

Initial investigations have been focused on multiscale topology optimization (MTO) methods, i.e., mapping relationships between pseudo-densities and macroscopic properties for a representative volume element (RVE), where macroscopic properties are obtained employing homogenization methods [15]. Trained models are used to predict macroscopic properties in real-time without executing the homogenization computation repeatedly. Watts et al. [16] employed specified truss RVEs, which are symmetric and isotropic. The mapping error between the pseudo-density and four properties (such as the axial Young's modulus) is controlled to 5%. Likewise, Xiao and Zhang et al. [17,18] modeled the mapping relationship based on the Kriging metamodel to effectively address the natural frequency problem, where the geometry of the RVE is explicitly expressed, even allowing the geometric topology to vary segmentally with respect to the pseudo-density. However, the mapping relationship is more complex for RVEs with anisotropy. For instance, Kim et al. [19] applied a deep neural network (DNN) to predict the macroscopic properties of fiber-reinforced RVEs, which can capture more complex feature variations in mapping relationships. In practice, the paradigms dealing with MTO methods are almost identical [20–22], and only the strategy for constructing mapping relations is different. On the other hand, MTO methods focus on mapping the properties of the RVEs, and the optimization at the macroscopic level still follows traditional approaches. Therefore, it is necessary to research an ML-based approach to address single-scale TO.

With the rapid development of DNNs, integration with TO has been realized in different manners [23].

Generative design, such as non-iterative optimization design [24], and diversity generative design [25], might be the most popular. Among them, the major purpose is to predict the optimization results in real-time based on given information (e.g., boundary conditions, and optimization parameters). This paradigm has been successfully applied to 2D [26–28] and 3D [29,30] compliance minimization, as well as a meta-material design [31]. In some investigations, a multi-stage process was executed to generate high-resolution optimization results without iterations [24,32]. Another major application is to improve the efficiency of the topology optimization process. One strategy is to predict the optimization results using the density field information from early optimization iterations, thus the remaining optimization iterations can be skipped [33,34]. Another strategy is to use NN models to replace the finite element analysis (FEA) [35] or sensitivity analysis [36,37], and the application of DNN to replace both FEA and sensitivity analysis has received attractive attention [38]. In addition, there are also a variety of ways to integrate ML/DNN into TO [39–41].

Importantly, note that most of the reported methods above are data-driven paradigms, for which the computational cost of data acquisition is prohibitive, especially for large-scale problems. In reality, compared with FEA and other traditional computational methods, some applications of DNNs to low-resolution problems might waste computational resources. Additionally, although the data-driven paradigm has been a great success, the inference of DNN still depends on the training data, thus it is difficult for them to predict with unseen data. Chi et al. [36] proposed an online training strategy that cuts the pseudo-density and sensitivity fields from the early stage of optimization as training data. This strategy not only weakens the data dependency but also reduces the size of the input. Zhang et al. [42] parameterized pseudo-densities as parameters of the DNNs, and updating parameters is equivalent to updating the pseudo-densities distribution. Unfortunately, scaling up the problem size increases the number of neurons, which has a much higher demand on the computing power of a device.

Recently Raissi et al. [43] proposed an alternative method, i.e., physics-informed neural networks (PINNs) to solve partial differential equations (PDEs), which encode the PDEs into the loss of a DNN and can solve forward and inverse problems using minimal data. For forward problems, the PINN models can be well-trained without data acquisition [44]. Since its development, PINNs have successfully addressed issues such as fluid flow [45,46], heat transfer [47], material identification [48], and solid mechanics [44,49]. According to the form of loss functions, a PINN can be divided into the energy-based approach [50,51] and the PDE-based approach [52,53]; for more details refer to [44]. Among them, solving the PDEs in solid mechanics by the energy-based approach shows a high prediction accuracy for given examples [51] and also has the advantage of few hyperparameters and excellent convergence. Therefore, a natural idea is to integrate energy-based PINNs into TO. However, there have been only a few related investigations, which have not been well explored, and some representative works are only briefly described. Jeong et al. [54,55] applied an energy-based PINN to successfully address TOs, where the internal energy is calculated at the collocations and the material properties are mapped from elements to collocations. Subsequently, a sensitivity-analysis PINN was proposed to update the design variables and cooperate with the PINN that was used to replace FEA; thus a complete PINN-based TO was achieved. He et al. [56] also applied an energy-based PINN that not only addressed the compliance minimization TO but also extended to metamaterial design TO. Here, the internal energy was calculated at elements, thus avoiding the mapping of materials. However, the above PINN-based TOs have not been extended to multi-load problems and self-adjoint problems.

Therefore, a novel dynamically configured physics-informed neural networks-based topology optimization (DCPINN-TO) method is proposed in this paper, in which the trainable parameters will be dynamically configured and aim to replace expensive NN with a cheap one. Compared to the FEA-based TO, the FEA-based evaluation is replaced by DCPINNs, and the FE modeling procedure is omitted. To further save training costs, an active sampling strategy is proposed to reduce the number of collocations. Subsequently, the authors realized that the PDEs for both the adjoint equation and the governing equation are identical, and only the boundary conditions are different when treating the right-hand side of the adjoint equation as a pseudo-loading. Therefore, the proposed method is extended to the compliance minimization problem with displacement constraints. The paper is organized as follows. In Section 2, the DCPINN-TO framework with the active sampling strategy is described in detail. In Section 3, several classic numerical examples are implemented to demonstrate the competency of the proposed method. Finally, the conclusion is given in Section 4.

## 2. DCPINN-TO

The dynamically configured physics-informed neural network-based topology optimization (DCPINN-TO) is proposed in this section. To begin with, a brief review of the topology optimization model is presented, in which displacement constraints are taken into account; Subsequently, the network architecture of the DCPINN and the method of calculating the internal energy are suggested; To conclude, the complete implementation process of the DCPINN-TO with an active sampling strategy is presented.

### 2.1 Topology optimization

The DCPINN-TO method in this paper is built on the popular SIMP method [4] but is likewise suitable for other TO approaches. The SIMP method has the advantage that discrete variables are converted to continuous variables (material interpolation), weakening the difficulty of optimization. Note that the material interpolation works equivalently to stress interpolation in the linear elastic mechanical system, which is expressed as follows,

$$E = (\eta + (1-\eta)\rho^p)E_0. \tag{1}$$

The notation $\eta$ is the ratio of Young's modulus, $\eta = E_{min}/E_0$, where $E_0$ corresponds to the solid elements and $E_{min}$ corresponds to the void elements. Void elements are assigned properties in the SIMP method to avoid singularities. The notation $p$ is a penalty factor designed to drive the pseudo-density, $\rho$, to converge to a sharp 0/1 solution. In this paper, the ratio $\eta$ is set to $10^{-6}$, and the penalty factor $p$ is set to 3. With material interpolation, a classical TO problem referred to as compliance minimization is applied to validate the feasibility of DCPINN-TO. In addition, its application scenarios are extended to multi-load and multi-constraint problems. The compliance minimization problem is described as follows:

$$\begin{aligned}
&\min: f_c(\boldsymbol{\rho}) \\
&s.t.: \nabla \cdot \boldsymbol{\sigma}_m(\boldsymbol{\rho}) = 0, \quad m = 1,2,... \\
&\quad\quad g_k(\boldsymbol{\rho}) \leq 0, \quad k = v,d \\
&\quad\quad 0 \leq \rho_e \leq 1, \quad e = 1,2,...,N
\end{aligned} \tag{2}$$

The notation $m$ is the number of loads (working conditions) and the notation $N$ is the number of the discrete elements. The notation $f_c(\boldsymbol{\rho})$ denotes the optimization objective named compliance, which can be expressed as a function of Cauchy stress $\sigma$ and Cauchy strain $\varepsilon$,

$$f_c(\boldsymbol{\rho}) = \frac{N}{2f_{c,0}} \int_\Omega \boldsymbol{\sigma}(\boldsymbol{\rho}) : \boldsymbol{\varepsilon} \, dv. \tag{3}$$

The notation $f_{c,0}$ denotes the initial compliance. Note that the Cauchy stress $\sigma$ is linearly related to Young's modulus and is expressed as follows,

$$\boldsymbol{\sigma}(\rho) = (\eta + (1-\eta)\rho^p)\boldsymbol{\sigma}_0. \tag{4}$$

The Cauchy stress $\sigma_0$ is calculated in solid elements. The notation $g_v(\boldsymbol{\rho})$ denotes the volume constraint with the predefined volume fraction $\bar{V}_0$, which is expressed as:

$$g_v(\boldsymbol{\rho}) = N(\int_\Omega \rho \, dv - \bar{V}_0), \tag{5}$$

Likewise, the notation $g_d(\boldsymbol{\rho})$ denotes the displacement constraint with the predefined displacement limit $\bar{u}_0$, , which is expressed as:

$$g_d(\boldsymbol{\rho}) = \frac{N}{f_{c,0}}(\bar{u} - \bar{u}_0). \tag{6}$$

The notation $u$ is the nodal displacement at the specified position. In practice, the integral operation is replaced by a discretized formulation, i.e., the finite element method (FEM). Therefore, the compliance and volume of each element are significantly correlated to their geometric size. To avoid the effect of mesh refinement on their gradients (optimization objective and constraints), the coefficients $N/f_{c,0}$ and $N$ are applied. The optimization objective and displacement constraint are displacement dependent, so the initial compliance $f_{c,0}$ is added to scale.

After discretization based on the FEM, the FE formats of the optimization objective and the volume constraint can be written as follows:

$$f_c(\boldsymbol{\rho}) = \frac{N}{2f_{c,0}} \int_\Omega \boldsymbol{\sigma}(\boldsymbol{\rho}) : \boldsymbol{\varepsilon}\, dv = \frac{N}{2f_{c,0}} \boldsymbol{u}^\mathrm{T} \mathbf{K}(\boldsymbol{\rho}) \boldsymbol{u}, \tag{7}$$

$$g_v(\boldsymbol{\rho}) = N\left(\left(\sum_{i=1}^N \rho_i v_i\right) - \overline{V}_0\right), \tag{8}$$

$$g_d(\boldsymbol{\rho}) = \frac{N}{f_{c,0}}(\boldsymbol{L}^\mathrm{T}\boldsymbol{u} - \overline{u}_0). \tag{9}$$

The notation $\mathbf{K}$ denotes the global stiffness matrix, the notation $v_i$ denotes the volume of the element $i$, and the notation $\boldsymbol{L}^\mathrm{T}$ is the one-hot vector with 1 at the specified position. Furthermore, the governing equation with load vector $\boldsymbol{t}$ can be written as follows:

$$\mathbf{K}\boldsymbol{u} = \boldsymbol{t} \tag{10}$$

According to the discrete expressions, the FE format of the topology optimization model (2) can be rewritten as follows:

$$\begin{aligned}
\min: &\ C(\boldsymbol{\rho}) = \frac{N}{f_{c,0}} \boldsymbol{u}^\mathrm{T}\mathbf{K}\boldsymbol{u} \\
\text{s.t.}: &\ \mathbf{K}_m \boldsymbol{u}_m = \boldsymbol{t}_m, \quad m = 1, 2, \ldots \\
&\ g_v(\boldsymbol{\rho}) = N\left(\left(\sum_{i=1}^N \rho_i v_i\right) - \overline{V}_0\right) \leq 0 \\
&\ g_d(\boldsymbol{\rho}) = \frac{N}{f_{c,0}}(\boldsymbol{L}^\mathrm{T}\boldsymbol{u} - \overline{u}_0) \leq 0 \\
&\ 0 \leq \rho_e \leq 1, \quad e = 1, 2, \ldots, N
\end{aligned} \tag{11}$$

Note that although the optimization model (11) is described by applying the FE format, the displacements used are the predictions of the DCPINN with the input $x$ and the trainable parameters $\boldsymbol{\theta}$, as follows:

$$\boldsymbol{u} = NN(\boldsymbol{x}; \boldsymbol{\theta}). \tag{12}$$

When the displacement constraint is not considered, the optimization model (11) degenerates into a classical compliance minimization problem.

The SIMP method is a gradient-based optimization method, so it is necessary to perform the sensitivity analysis of both the optimization objectives and constraints relative to the pseudo-density $\boldsymbol{\rho}$. Since TO commonly employs filtering techniques to avoid checkerboard phenomena, the actual physical density $\tilde{\boldsymbol{\rho}}$ used for FEA. In this section, only the partial derivatives relative to the physical density $\tilde{\boldsymbol{\rho}}$ are calculated, and a complete sensitivity analysis using the chain rule is given in Section 2.3. With the adjoint method, the partial derivative of the optimization objective at the element $e$ is written as follows,

$$\frac{\partial f_c(\boldsymbol{\rho})}{\partial \tilde{\rho}_e} = \frac{N}{2 f_{c,0}} \frac{\partial \boldsymbol{u}^\mathrm{T} \mathbf{K} \boldsymbol{u}}{\partial \tilde{\rho}_e} = -\frac{N(p - p\eta)}{2 f_{c,0}} \tilde{\rho}_e^{p-1} \boldsymbol{u}_e^\mathrm{T} \boldsymbol{k}_e^0 \boldsymbol{u}_e. \tag{13}$$

The subscript $e$ indicates association with element $e$. The partial derivative of the volume constraint at the element $e$ is directly calculated as:

$$\frac{\partial g_v}{\partial \tilde{\rho}_e} = \frac{\partial (N(\sum_{i=1}^{N} \tilde{\rho}_i v_i - V_0))}{\partial \tilde{\rho}_e} = N v_e. \tag{14}$$

Likewise, the partial derivative of the objective function at the element $e$ with the adjoint method is written as,

$$\frac{\partial g_d}{\partial \tilde{\rho}_e} = -\frac{N}{f_{c,0}} \boldsymbol{L}^\mathrm{T} \mathbf{K}^{-1} \frac{\partial \mathbf{K}}{\partial \tilde{\rho}_e} \boldsymbol{u}. \tag{15}$$

Assuming there is an adjoint vector $\boldsymbol{\lambda}$ that satisfies,

$$\mathbf{K}\boldsymbol{\lambda} = \boldsymbol{L}. \tag{16}$$

The equation (15) can be rewritten as:

$$\frac{\partial g_d}{\partial \tilde{\rho}_e} = -\frac{N}{f_{c,0}} \boldsymbol{\lambda}^\mathrm{T} \frac{\partial \mathbf{K}}{\partial \tilde{\rho}_e} \boldsymbol{u} = -\frac{N(p - p\eta)}{f_{c,0}} \tilde{\rho}_e^{p-1} \boldsymbol{\lambda}_e^\mathrm{T} \boldsymbol{k}_e^0 \boldsymbol{u}_e. \tag{17}$$

The PDE corresponding to the equation (16) is almost identical to the governing equation, only the Newman boundary condition is different when $\boldsymbol{L}$ is treated as a pseudo-load. Thus the DCPINN can also be employed to solve the following adjoint equation:

$$\boldsymbol{\sigma} \boldsymbol{n} = \boldsymbol{L}, \quad \forall \boldsymbol{x} \in \Gamma_{\tilde{N}}. \tag{18}$$

## 2.2 Energy-based DCPINN

For a homogeneous, isotropic, linear elastic mechanical system with the small deformation assumption, the governing equation can be written as:

$$\nabla \cdot \boldsymbol{\sigma} + \boldsymbol{f} = 0, \quad \forall \boldsymbol{x} \in \Omega. \tag{19}$$

The notation $\nabla$ denotes the gradient operator, and the notation $\boldsymbol{f}$ is the body force. The Cauchy stress $\boldsymbol{\sigma}$ can be calculated by the constitutive law as follows:

$$\boldsymbol{\sigma} = \frac{E_0 v}{(1+v)(1-2v)} tr(\boldsymbol{\varepsilon})\mathbf{I} + \frac{E_0}{(1+v)} \boldsymbol{\varepsilon}, \tag{20}$$

$$\boldsymbol{\varepsilon} = \frac{1}{2}(\nabla \boldsymbol{u}^\mathrm{T} + \nabla \boldsymbol{u}). \tag{21}$$

The notation $v$ denotes the Poisson's ratio. From the statement of the principle of virtual work [44], it is well-known that the virtual displacements represent all the possible configurations and only one configuration corresponding to the governing equation that makes the virtual work $\delta \Pi$ vanish, and this configuration is the real displacement $u$, as follows:

$$\delta \Pi = \delta U + \delta V. \tag{22}$$

The notation $\delta$ denotes the variation operator, and the notations $\delta U$ and $\delta V$ denote the internal virtual work and external virtual work, respectively. Assuming the absence of body force and that the Dirichlet and Neumann boundary conditions are satisfied,

$$u = \bar{u}, \quad \forall x \in \Gamma_D, \tag{23}$$
$$\sigma n = \bar{t}, \quad \forall x \in \Gamma_N, \tag{24}$$

where $\Gamma_D$ and $\Gamma_N$ denote the Dirichlet and Neumann boundaries, respectively, $\bar{u}$ is predefined displacement, and $\bar{t}$ is the external tractive force. Associating the above equations, the potential energy of the system can be rewritten as:

$$\delta \Pi = \frac{1}{2} \int_\Omega \sigma : \varepsilon \, dv - \int_{\Gamma_N} \bar{t} \cdot u \, ds. \tag{25}$$

The key to an energy-based PINN is to employ the potential energy as a loss function to search for the true displacement solution by finding the stationary value (minimum potential energy) of the loss function.

The architecture of the proposed DCPINN model consists of two sub-networks, namely, backbone neural network (NN) and coefficient NN, as shown in Figure 1. The backbone NN is a fully connected feed-forward network, which is composed of an input layer, three hidden layers, and an output layer, as follows:

$$\tilde{u} = NN_{bac}(\chi; \theta_1) = f_n^L \circ \phi \circ f_{n-1}^L \circ \phi \circ ... \phi \circ f_1^{FT}(\chi);$$
$$f_l^L(a) = W_l a + b_l, \quad l = 2,...,n. \tag{26}$$

The notation $\phi$ is the activation function, the notation $f_l^L(\cdot)$ is a linear layer, and the notation $\theta_1$ are the trainable parameters including $W_l$ and $b_l$, which are the parameters between layers. To improve the accuracy of the backbone NN, the Fourier transformation [57] is used as the input layer, $f_1^{FT}(\cdot)$, to transform input features to the frequency domain. For the backbone NN, the input features are coordinates $\chi = [x, y, z]^T$, and the output features are nodal displacements $\tilde{u} = [\tilde{u}_x, \tilde{u}_y, \tilde{u}_z]^T$. The number of neurons in the input and output layers depends on the dimension of the given problem, and each hidden layer has 360 neurons.

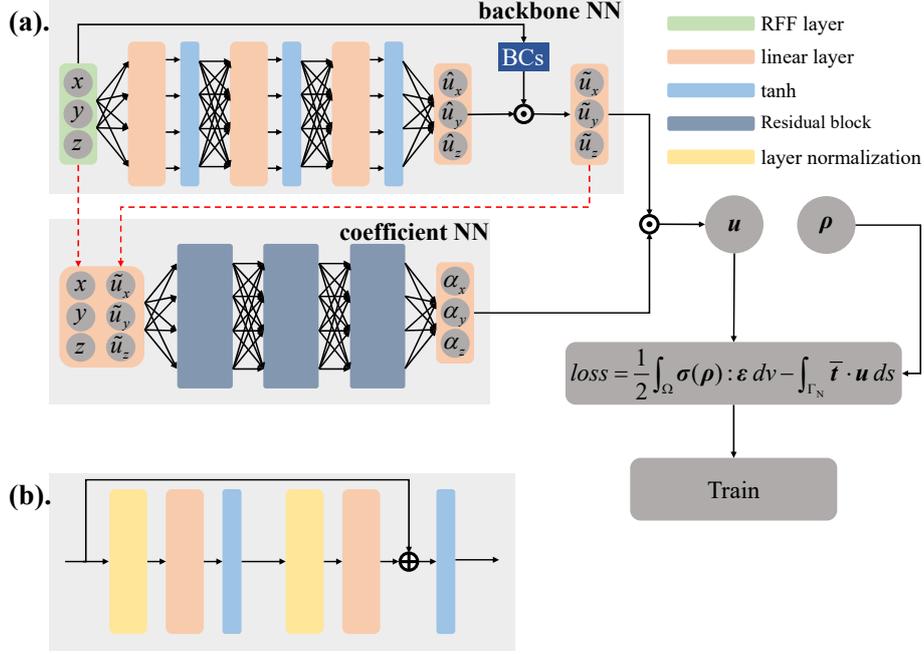

Figure 1 Schematic diagram of energy-based DCPINN: (a) architecture of the DCPINN and (b) Residual block in the coefficient NN.

Instead of a fully connected feed-forward network, the coefficient NN uses the Residual block ($f_l^R(\cdot)$, and is presented in Figure 1 (b)) as the hidden layers. The Residual block comprises two sequential stacks of normalized (layer normalization) $f^N$, and linear $f^L$ layers. Each linear layer of the Residual block has 48 neurons. Notably, the input of the coefficient NN consists of the outputs of the backbone NN and the coordinates $[\tilde{u}_x, \tilde{u}_y, \tilde{u}_z, x, y, z]^T$ as follows:

$$\begin{aligned} \boldsymbol{\alpha} &= NN_{coe}(\tilde{\boldsymbol{u}}, \boldsymbol{\chi}; \boldsymbol{\theta}_2) = f_n \circ \phi \circ f_{n-1}^R \circ \phi \circ ... \phi \circ f_2^R \circ f_1(\tilde{\boldsymbol{u}}, \boldsymbol{\chi}); \\ f_l^R(\boldsymbol{a}) &= \phi \circ (\boldsymbol{a} + f^L \circ f^N \circ \phi \circ f^L \circ f^N(\boldsymbol{a})), \quad l = 2, ..., n-1. \end{aligned} \quad (27)$$

The final prediction, $\boldsymbol{u} = [u_x, u_y, u_z]^T$, is obtained by multiplying the output of the coefficient NN $\boldsymbol{\alpha}$ by the corresponding output of the backbone NN $\tilde{\boldsymbol{u}}$. In addition, the activation function tanh is used in both NNs.

$$\begin{cases} \tilde{\boldsymbol{u}} = NN_{bac}(\boldsymbol{\chi}; \boldsymbol{\theta}_1) \\ \boldsymbol{\alpha} = NN_{coe}(\tilde{\boldsymbol{u}}, \boldsymbol{\chi}; \boldsymbol{\theta}_2) \\ u_i = \alpha_i \times \tilde{u}_i \quad , i = x, y, z \end{cases} \quad (28)$$

Note that the hard constraint [58] is applied to act on the features $\boldsymbol{u}$ to satisfy the Dirichlet boundary conditions, The implementation of the hard constraint relies on two auxiliary functions as follows:

$$\boldsymbol{u} = g(\boldsymbol{x}) + l(\boldsymbol{x})\tilde{\boldsymbol{u}}, \quad (29)$$

where the role of the auxiliary function $g(\boldsymbol{x})$ is to satisfy the Dirichlet boundary condition and the role of the auxiliary function $l(\boldsymbol{x})$ is to impose the sample points at the Dirichlet boundary to be zero as follows:

$$\begin{cases} g(\boldsymbol{x}) = \bar{\boldsymbol{u}} \text{ and } l(\boldsymbol{x}) = 0, & \boldsymbol{x} \in \Gamma_D \\ g(\boldsymbol{x}) = 0 \text{ and } l(\boldsymbol{x}) > 0, & \boldsymbol{x} \notin \Gamma_D \end{cases}. \qquad (30)$$

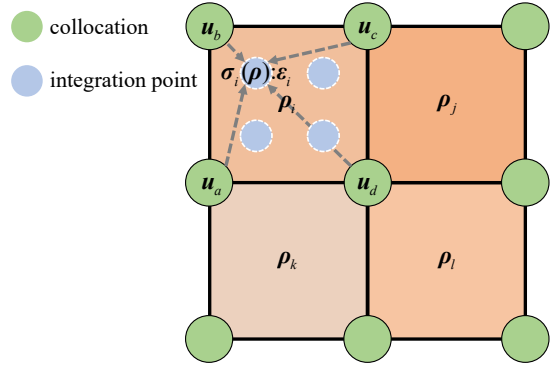

Figure 2 Calculation of the strain energy density employing Gaussian integration.

During the iterative process of topology optimization, Young's modulus is interpolated by the pseudo-density; the pseudo-density field is non-uniform with a high contrast. However, the pseudo-density is an inherent property of each element, and the mapping between elements and collocations can affect the accuracy of the interpolation of Young's modulus. Jeong et al. [54] determined the pseudo-density at the collocation using the pooling method, These researchers revealed that the pseudo-density mapping for either maximum pooling or average pooling was inaccurate. Accurate material properties are crucial for accurately calculating internal energy. For this reason, Gaussian integration is employed to calculate the internal energy of the elements instead of collocations (Figure 2), which yields a by-product of decoupling the material from the stress/strain. The stress/strain does not have to be calculated at collocations.

$$\int_{\Omega_e} \boldsymbol{\sigma}_e(\boldsymbol{\rho}) : \boldsymbol{\varepsilon}_e dv = \sum \kappa_i \boldsymbol{\sigma}_i(\boldsymbol{\rho}) : \boldsymbol{\varepsilon}_i, \qquad (31)$$

with

$$\boldsymbol{\sigma}_i(\boldsymbol{\rho}) = \psi(\boldsymbol{\rho}) \boldsymbol{\sigma}_{i,0}. \qquad (32)$$

The notation $\psi(\boldsymbol{\rho})$ denotes the material interpolation which was introduced in the previous section, and the notation $\kappa_i$ is the determinant of the Jacobi matrix. The Cauchy stresses $\boldsymbol{\sigma}_{i,0}$ and strains $\boldsymbol{\varepsilon}_i$ are associated with pure solids, i.e., $\rho=1$. The notation $i$ denotes the index of the Gaussian integration point. More critically, all integration points are within the element, so the exact pseudo-densities of the element can be used directly to interpolate the stress/strain. Therefore, it can be guaranteed that the utilization of Young's modulus for the calculation of internal energy enables acceptable accuracy. Note that $\Omega_e$ is the elemental domain, which is different from the overall structural domain $\Omega$ in equation (25).

## 2.3 The process of DCPINN-TO

The DCPINN built-in DCPINN-TO can predict the displacement field without observed data, which can replace two important modules in each optimization cycle: the assemblage of stiffness matrice, and FEA. Intuitively, compared with a high pseudo-density element, a low pseudo-density element provides less internal energy, but has an equivalent training cost. For this reason, an active sampling strategy is suggested for selectively sampling at the high pseudo-density elements, which in turn reduces computational costs. In addition, the pseudo-density distribution between successive cycles is approximate; consequently, the previously trained DCPINN model can be used as a pre-trained model for the current optimization cycle. Note that the trainable parameters are dynamically configured to further save costs considering the similarity of the pseudo-density distribution. Therefore, a coefficient NN is more inexpensive to train and is designed to replace the backbone NN at certain cycles, thus the training costs are further economized. The process of DCPINN-TO is presented in Figure 3 for comprehension.

To begin with, the pseudo-densities are uniformly distributed in the design domain with a value of the predefined volume limit $\overline{V}_0$. As the optimization proceeds, the pseudo-densities are driven to converge to a sharp 0/1 solution due to the penalty effect attributed to the material interpolation (1). However, the void elements, i.e., $\rho = 0$, represent the absence of solid material in the physical sense and therefore cannot provide internal energy. Similarly, the internal energy provided by an element with a pseudo-density close to zero is minuscule with respect to the whole system. Although all the collocations are used to train the DCPINN models, the high contrast pseudo-density field results in differences in the contribution in terms of internal energy. Whereas during training, attention is more likely to be attracted to elements that provide more internal energy, and the predictions of the collocations associated with them will be more accurate. Clearly, training collocations associated with low pseudo-densities have the same cost but do not achieve the same prediction accuracy as those with high pseudo-densities. A useful conclusion by Zheng et al. [59] reveals that updating design variables extremely close to 0 is unnecessary. For this reason, an active sampling strategy coupled with the Heaviside projection [60] is proposed to reduce the training cost of the DCPINN; i.e., only the collocations associated with physical density variables of more than a threshold value $\tau$ are employed for training.

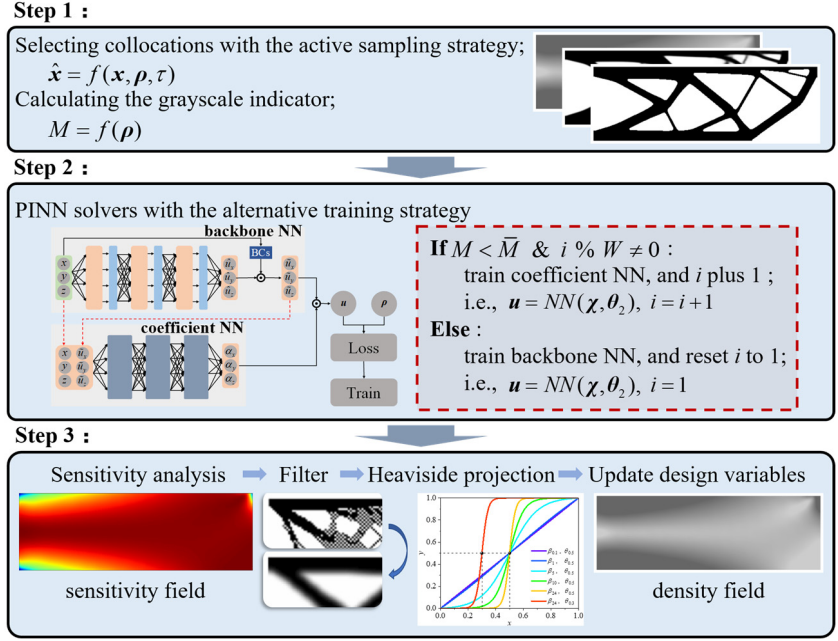

Figure 3 Process of DCPINN-TO with active sampling strategy.

$$\hat{x}(\rho) = \{x \in \Omega \mid \tilde{\rho}_e > \tau,\ e = 1, 2, ..., N\}, \tag{33}$$

$$\tilde{\tilde{\rho}}_i = \frac{\tanh(\beta\theta_h) + \tanh(\beta(\tilde{\rho}_i - \theta_h))}{\tanh(\beta\theta_h) + \tanh(\beta(1 - \theta_h))}. \tag{34}$$

The notation $\tilde{\rho}$ is the filter density, the notation $\theta_h$ is the threshold of the Heaviside projection, and the notation $\beta$ controls the curvature (as shown in Figure 4). In practice, $\beta$ is doubled every 5 optimization cycles until it reaches a predefined maximum value. The threshold $\theta_h$ is determined by a one-dimensional search, such as a bisection. Note that the filter density $\tilde{\rho}$ is obtained by density filtering as follows:

$$\tilde{\rho}_i = \frac{\sum_{j \in N_i} w_i(x_j)\rho_j}{\sum_{j \in N_i} w_i(x_j)}, \tag{35}$$

$$w_i(x_j) = \max(0,\ r - \|x_j - x_i\|). \tag{36}$$

Subsequently, the DCPINN models are employed to predict the displacement solutions and the adjoint solutions, and this process is key to the successful implementation of DCPINN-TO. As mentioned in the above section, the network architecture consists of a backbone NN and a coefficient NN. However, training both networks simultaneously is unnecessary; therefore, a grayscale indicator $M$ is introduced to manage the training strategy by dynamically configuring the trainable parameters:

$$M = \frac{4\sum_{i=1}^{N}\tilde{\rho}_i(1-\tilde{\rho}_i)}{N}. \tag{37}$$

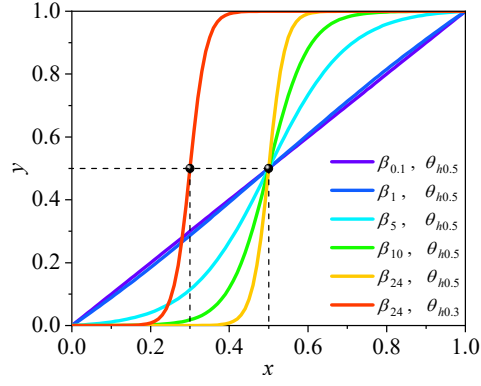

Figure 4 Heaviside projection with different parameters.

If the grayscale $M$ is more than the predefined grayscale limit $\bar{M}$, then only the backbone NN is trained while all coefficients $\alpha_i$ are fixed as 1. If the grayscale $M$ is less than the predefined grayscale limit $\bar{M}$, then the optimization process is considered to enter a fine-tuned state; i.e., the changes in the pseudo-density field are small. Thus in this stage, $N_{op}$ optimization cycles are treated as one period. For the first optimization cycle of a period, only the backbone NN is trained, and for the other optimization cycles only the coefficient NN is trained. The Adam algorithm [61] is a common optimizer in ML; therefore, it is also used in this paper. One thousand epochs are needed when only the coefficient NN is trained, and 3000 epochs are needed for other situations. Given that the pseudo-density distribution of successive iterations is close, and inspired by the concept of migration learning, the previously trained model parameters are used as the pre-trained parameters of the current DCPINN model. In addition, although the PDEs for the governing equations and the adjoint equations are almost identical except for the Newman boundary, the governing equations and the adjoint equations corresponding to different DCPINN models are suggested in this paper.

Then, sensitivity analysis is necessary for the gradient-based SIMP method. To improve the robustness of DCPINN-TO, a sensitivity filtering technique is employed to smooth the sensitivity distribution as follows:

$$\frac{\partial (\cdot)}{\partial \tilde{\rho}_i} = \frac{\sum_{j \in N_i} w_i(\boldsymbol{x}_j) \rho_j \frac{\partial (\cdot)}{\partial \tilde{\tilde{\rho}}_i}}{\rho_i \sum_{j \in N_i} w_i(\boldsymbol{x}_j)}, \qquad (38)$$

where $(\cdot)$ can be either the objective or the constraint functions. After the sensitivity filtering, the sensitivity of the objective and constraint functions relative to the pseudo-density $\boldsymbol{\rho}$ are calculated by the chain rule.

$$\frac{\partial (\cdot)}{\partial \rho_e} = \sum_{i \in N_e} \frac{\partial (\cdot)}{\partial \tilde{\tilde{\rho}}_i} \frac{\partial \tilde{\tilde{\rho}}_i}{\partial \tilde{\rho}_i} \frac{\partial \tilde{\rho}_i}{\partial \rho_e}, \qquad (39)$$

$$\frac{\partial \tilde{\tilde{\rho}}_i}{\partial \tilde{\rho}_i} = \beta \frac{1-\tanh(\beta(1-\theta))^2}{\tanh(\beta\theta)+\tanh(\beta(1-\theta))}, \tag{40}$$

$$\frac{\partial \tilde{\rho}_i}{\partial \rho_e} = \frac{w_i(\boldsymbol{x}_e)}{\sum_{j \in N_i} w_i(\boldsymbol{x}_j)}. \tag{41}$$

Eventually, the stopping criterion is used to control the exit of the optimization, otherwise returns to the active sampling step. The stopping criterion can be set to a fluctuation threshold of the density [62] or the optimization objective [5]. In this paper, the stopping criterion is described as follows:

$$\tau_{stop} = \frac{\sum_{i=0}^{Ns-1} |f_c^{(k-i)}(\boldsymbol{\rho}) - f_c^{(k-i-Ns)}f(\boldsymbol{\rho})|}{\sum_{i=0}^{Ns-1} f_c^{(k-i)}(\boldsymbol{\rho})}, \tag{42}$$

where $Ns$ is an integral number, usually 3.

## 3. Example results and discussion

The competency of the proposed DCPINN-TO method is verified by several examples. If not otherwise specified, then Young's modulus is set to 210 kPa, Poisson's ratio is set to 0.3, the parameter $\beta$ increases from 0.1 to a maximum of 24, the grayscale limit $\bar{M}$ is 5%, the value $N_{op}$ is 3, the stopping criterion is $10^{-4}$, and the threshold value $\tau$ in the Heaviside projection is $10^{-3}$. Furthermore, for all the numerical experiments, the computer system is equipped with an NVIDIA GeForce RTX 4060 Ti 8G processor with 16G of RAM.

### 3.1 Compliance minimization problem with different resolution

In this section, two examples are employed to verify the feasibility and competence of the proposed DCPINN-TO method. As presented in Figure 5, the boundary conditions and the initial geometric domain are different for the two examples. The left side of the cantilever beam is fixed and the right side is subjected to a static load. The upper side of the L-shaped beam is fixed and the right side is subjected to a static load. In both examples, the static loads (2 kN) are applied to a horizontal boundary of 0.25 m, and the predefined volume fractions $\bar{V}_0$ are 0.4. The geometries of both examples are meshed into low- and high-resolutions. The filter radii are 2.5 times the element size in low-resolution models and 4 times the element size in high-resolution models.

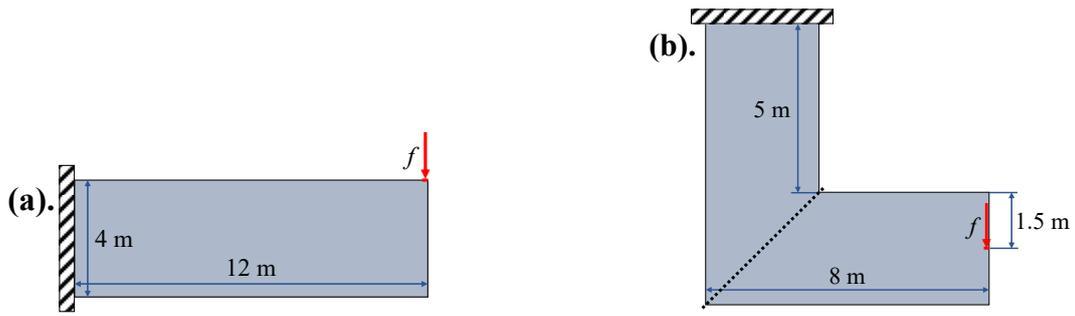

Figure 5 Design domains and boundary conditions for two examples: (a) is a 12-by-4 m2 cantilever beam, and (b) is an L-shape beam with characteristic dimensions of 8 m and 5 m. s

Note that the radii of high-resolution models are smaller on the physical scale; e.g., the filter radius of the cantilever beam is 0.0625 m for the low-resolution model, and 0.04 m for the high-resolution model. During the optimization process, the geometric domain (connected domain composed of non-void elements) is different for each optimization cycle, and the material field is inhomogeneous and high-contrast due to material interpolation. Nonetheless, DCPINN-TO demonstrates excellent availability. The optimized structures are presented in Figure 6, in which the components of the structures are connected normally and the cross-section of each component changes smoothly. There are no "islands" and disconnections for pseudo-density distribution, which may be found in other neural network-based topology optimization methods [24,27]. Even when dealing with high-resolution models, DCPINN-TO is also competitive, and the pseudo-density distribution approximates that of low-resolution models. Naturally, more detailed structures exist in the high-resolution models due to the reduced radii. Apart from that, the gradient of the objective function depends on the elemental strain energy, but its value vanishes with decreasing element size; this slows down the convergence of the optimization. Benefiting from the application of scalar coefficients, the effect of mesh refinement on the gradients is significantly weakened. In addition, Heaviside projection is used to accelerate the density field convergence to a sharp 0/1 solution. In the given examples, it usually takes dozens of iterations to meet the stopping criterion.

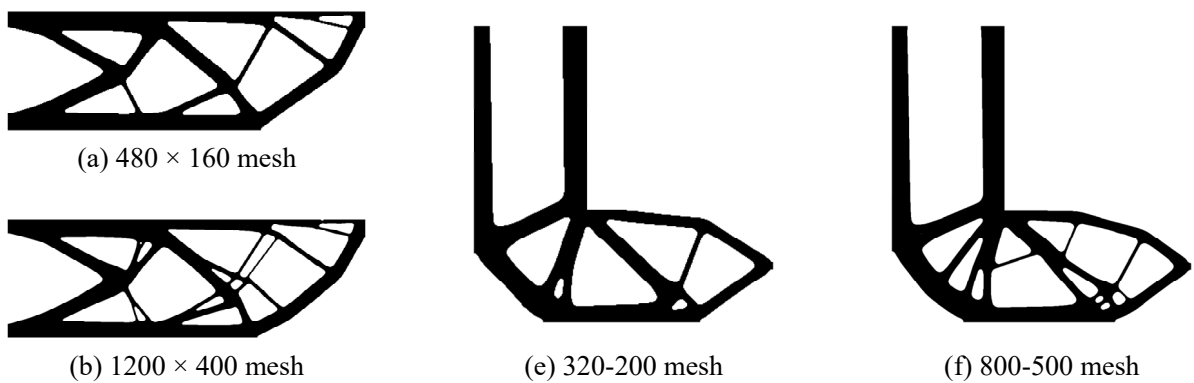

(a) 480 × 160 mesh

(b) 1200 × 400 mesh          (e) 320-200 mesh          (f) 800-500 mesh

Figure 6 The optimized structures for both examples at different resolutions.

The proposed DCPINN-TO is successfully applied in compliance minimization with different resolutions. In addition to the optimized structures, the effect of dynamically configuring trainable parameters and the active sampling strategy during the optimization process is also discussed in this section. The relationship between the grayscale and the collocations ratio of the cantilever beam is used to explain the role of the active sampling strategy. In Figure 7, the trends for both low- and high-resolution models are nearly identical. The curves of both the grayscale and the collection ratio gradually decrease as the optimization proceeds, because the density field tends to be a sharp 0/1 solution. The grayscale can converge to near 0, but the collocations ratio is dominated by the volume fraction. This phenomenon also explains why grayscale is applied as a switch for configuring trainable parameters rather than the collection ratio. In addition, both the grayscale and the collection ratio exhibit a step change, which is caused by the change in the parameter $\beta$ of the Heaviside projection. Bounded by the grayscale limit $\bar{M}$, TO can be divided into two stages; the high-gray stage and the low-gray stage. Note that there is no change in the collocations ratio at the beginning while grayscale decreases rapidly, but the training costs are directly related to the collocations ratio.

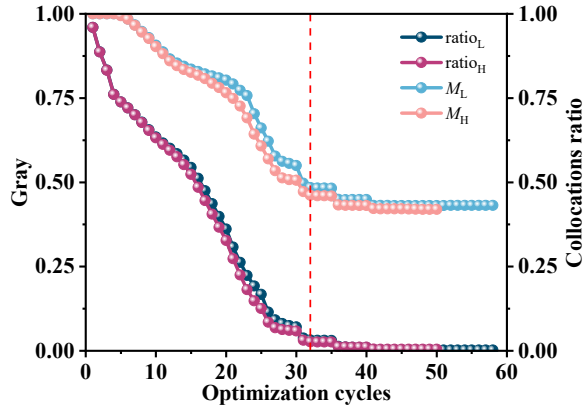

Figure 7 The grayscale and collocations rate during the optimization for the cantilever beam.

The curves containing historical information regarding the optimization process are presented in Figure 9 and Figure 8. At the high-gray stage, only the backbone NN needs to be merged into training, and the outputs of the coefficient NN $\lambda_i$ are all fixed as 1. As the TO proceeds, the material distribution gradually converges to a sharp 0/1 solution, so the reduction in trainable collocations naturally results in a reduction in the training costs of the DCPINN models. According to the statistics of both examples during the high-gray stage, the minimum training cost is close to 55% of the maximum training costs at low-resolution resolutions and 49% at high-resolution. The difference in training costs is due to the presence of different resolutions, the higher the resolution is, the more significant the cost savings. Figure 7 explains this phenomenon, the collocations ratio is lower in the high-resolution model than it is in the low-resolution model after more than a dozen optimization cycles. The active sampling strategy effectively reduces low-contribution collocations, which in turn improves the efficiency of model training.

Although this strategy may cause errors, the optimization results prove that the errors are acceptable, at least for the optimization process. A detailed error analysis will be discussed in Section 3.2.2 and Section 3.3.

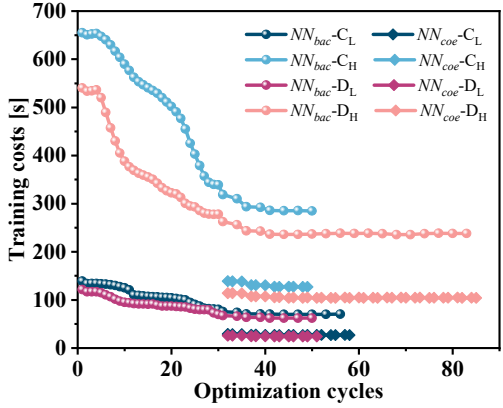

Figure 8 The training costs during the optimization for all examples.

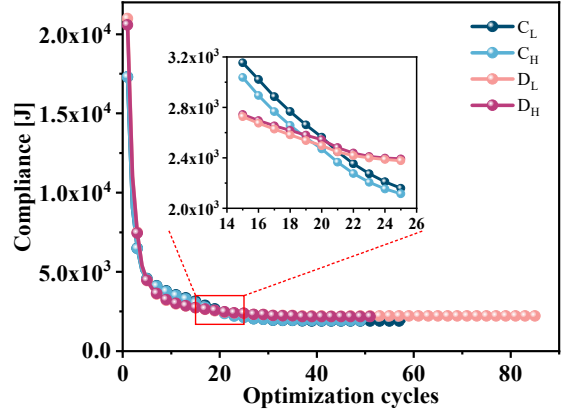

Figure 9 The optimization objectives during the optimization for all examples.

At the low-gray stage, the pseudo-density is very slightly modified (presented in Figure 10), and it can be inferred that the displacement field is likewise slightly modified provided that the boundary conditions are constant. For the DCPINN, the inexpensive coefficient NN is trained and the backbone NN is fixed in the current scenario as a feasible scheme. This approach avoids the accumulation of changes in the material distribution, which leads to a decline in the prediction accuracy of the DCPINN model. For each period ($N_{op}$ optimization cycles are treated as one period), only the coefficient NN is trained except for the last cycle where only the backbone NN is trained. In practice, the displacements by the backbone NN can be regarded as a series of basis terms, which are then interpolated by the outputs of the coefficient NN, thus obtaining the feasible displacements. Despite the displacement of each degree of freedom (DOF) being linearly interpolated, the entire displacement field is nonlinearly interpolated because of the nonlinear coefficient field. Note that this paper only presents a conservative network architecture for robustness. A more aggressive network architecture can be applied to further reduce training costs. Although the network architecture is conservative, the advantages of alternative training are obvious, significantly decreasing training costs while maintaining prediction accuracy. The costs of training the coefficient NN save approximately 62% at low-resolution resolutions and 56% at high-resolution resolutions, compared to the basis network.

In addition, the curves of the optimization objective (compliance) vary smoothly at different resolutions and almost overlap. This result verifies that the DCPINN-TO method possesses the same advantages as the FE-based topology optimization method; that is, mesh refinement hardly affects the optimization results, except by adding more details. The more details in high-resolution models are due to the change in the filter radius on the physical scale; however, the overall trend of the material distribution remains consistent. Additionally, the total iterations for topology optimization under the stopping criterion of $10^{-4}$ are only a few dozen steps, both for different geometric models and resolutions; This may indicate the good convergence of the proposed DCPINN-TO method. The test results in this section verify that the proposed DCPINN-TO method is feasible and efficient.

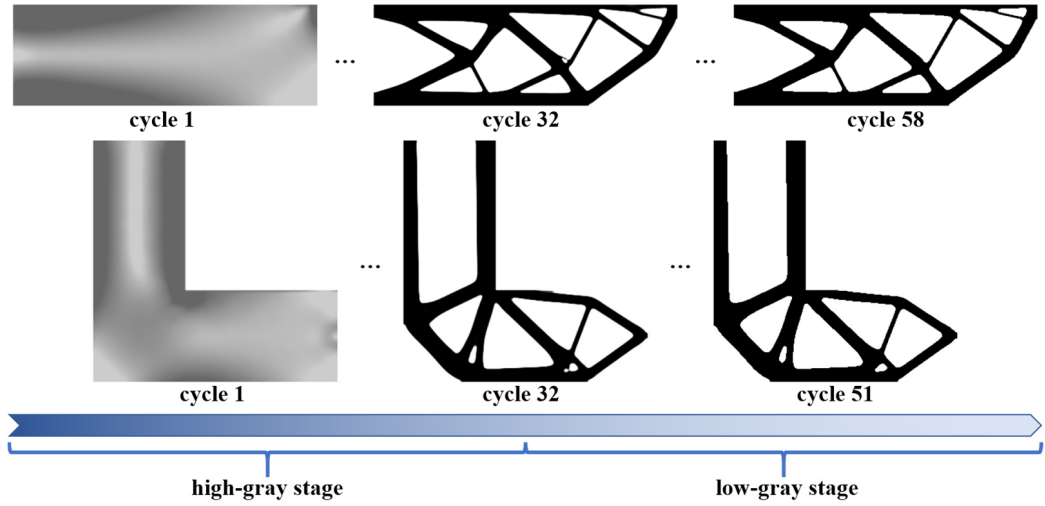

Figure 10 The variations of the pseudo-density field.

## 3.2 Multi-load and multi-constraint problems

The well-known compliance minimization problem is categorized as a self-adjoint problem because the adjoint equation is the governing equation itself. To expand the application scenarios, the multi-load problem consisting of multiple working conditions, and the multi-constraint problem consisting of volume constraints and displacement constraints are discussed in this section. The former remains a self-adjoint problem despite the presence of multiple working conditions (multiple loads), but the latter is an adjoint problem due to the volume constraint.

### 3.2.1 Multi-load problem with different volume fractions

The multi-load example is a 15-by-3 $m^2$ double-clamped beam with both circular and square holes, which is discretized to an 800 × 160 mesh. A total of two loads are applied to the structure. The first load $f_1$ is a 2 kN force with a 0.25 m boundary, which is applied to the center. The second load $f_2$ includes two 2 kN forces with a 0.125 m boundary, which are applied to the upper boundary at a distance of 1/4 from each end. The remaining parameters are consistent with Section 3.1 except that the volume fractions are 0.3 and 0.5, and the filter radius is 3 times the element size.

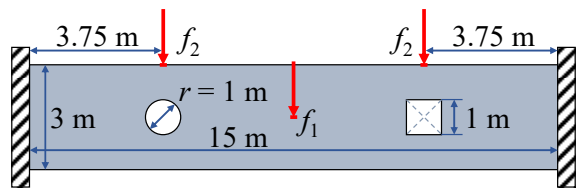

Figure 11 Design domains and boundary conditions for multi-load examples.

The optimized structures with different volume fractions are presented in Figure 12. To begin with, the irregular design domain does not seem to affect the optimization process, and elegant results can still be produced by the DCPINN-TO method. Despite the holes located at the main force transfer path, the pseudo-density distribution bypasses the holes and improves the carrying capacity of the structure as much as possible. The pseudo-density is enriched at the location where the loads are applied, which is consistent with the intuitive results. Subsequently, both optimized structures are asymmetric. Due to the differences in the shape of the holes, this phenomenon reflects that the DCPINN-TO method can capture and adapt to the complex geometric features of the design domain. Notably, each load corresponds to a separate DCPINN model. Otherwise, the errors in gradients (discussed in the next section) are superimposed because the gradients used to update density are related to the outputs of the DCPINN model. However, the elegant results reveal that the interference of errors in the optimization process can be negligible. Finally, the volume fraction can affect the pseudo-density distribution, but the main trend of the structural configuration is not affected by the volume fraction. The difference is only in some regions far from the bearing, probably because these regions have the least influence on the carrying capacity.

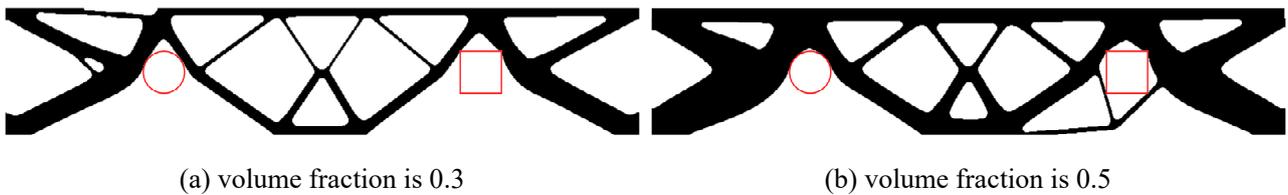

(a) volume fraction is 0.3    (b) volume fraction is 0.5

Figure 12 Optimized structures for the multi-load examples.

As presented in Figure 13, the change in compliance remains smooth and eventually stabilizes with a few dozen iterations. Differences in volume fractions are responsible for the inconsistency of the compliance curves. Clearly, the more material there is, the more the structure resists deformation, and the smaller compliance. There is insufficient evidence to prove that there is an essential correlation between the number of optimization cycles and the volume fraction, and the needed number of optimization cycles can be observed to be close in the given examples. The robustness of DCPINN-TO in terms of volume fraction further demonstrates its good scalability.

The simple illustration in Figure 7 shows the variation in the number of trainable collocations, and in this section, a more detailed comparison reveals the effect of volume fractions on trainable collocations with the active sampling strategy. As presented in Figure 14 (b), the number of trainable collocations decreases more rapidly in optimizations with small volume fractions and eventually stabilizes at a position slightly larger than the volume fraction (the collocations ratio, 33.20% for a volume fraction of 0.3, and 52.72% for a volume fraction of 0.5). Smaller volume fractions drive more elements to converge to a pseudo-density below the threshold during the optimization. Thus, trainable collocations decrease faster, and the gap widens further as the optimization proceeds until it stabilizes. The reason that the final collocations ratio is slightly higher than the volume fraction is due to the difficulty of eliminating the intermediate pseudo-density elements at the boundaries, which is inherent to the nature of the filtering technique. The reductions in trainable collocations are reflected in training costs, and an approximate decreasing trend is observed. In addition, the training costs for both the governing equations corresponding to different loads almost overlap, since the same collocations are used for both during each optimization cycle. The strategy of dynamically configuring trainable parameters continues to play an important role, further saving more training costs. This strategy acts on both governing equations, and the cost savings are double that of the single-load problems. On the other hand, In contrast, the training cost of the coefficient NN as a proportion of the backbone NN is almost unaffected by the volume fraction, because the sample size of the network inputs is consistent. In practice, it is the architecture of both NNs and the training parameters, such as depth, width, and epoch, that affect the proportion of the training cost.

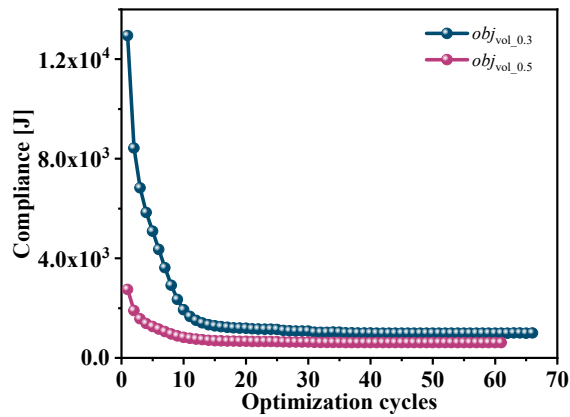

Figure 13 The compliance curves for the multi-load problem.

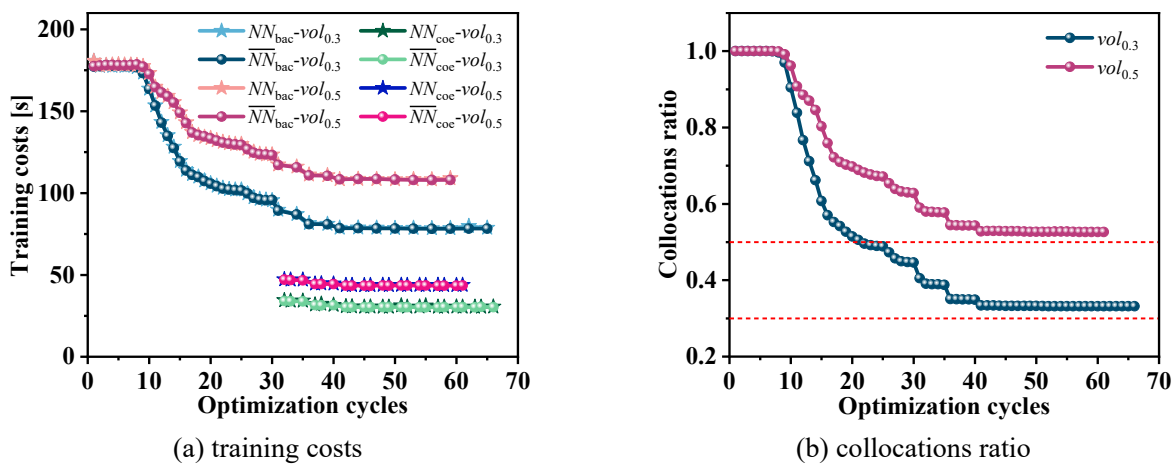

(a) training costs  (b) collocations ratio

Figure 14 Iterative curves for the multi-load problem, the upper horizontal lines of the legend in (a) are used to distinguish between different working conditions.

**3.2.2 multi-constraint problems and accuracy assessment**

It is well known that the compliance minimization problem with only a volume constraint is easy to solve. To extend the application scenarios of the DCPINN-TO method, an additional displacement constraint is integrated into the optimization model (as presented in Figure 15). The design domain is discretized to a 1000 × 200 mesh. The load is a 1 kN force with a 0.2 m boundary, applied to the center. The displacement constraint is applied to the midpoint of the bottom, which requires that the deformation of DOF in the vertical direction does not exceed 0.1 m. The remaining parameters are consistent with Section 3.1 except that the volume fractions are 0.3, and the filter radius is 3.5 times the element size.

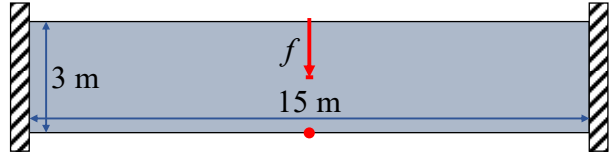

Figure 15 Design domains and boundary conditions for multi-constraint example.

In practice, imposing a strict displacement constraint at the initial stage is not recommended because of the exaggerated deformation at the current stage. Exaggerated differences between actual displacements and constraints cause the optimization to break down. For this purpose, a relaxed strategy is used in this example, i.e., an asymptotic displacement constraint, as follows:

$$\bar{u}_k^* = \max(0.9 u_{k-1}^*, \bar{u}_0), \tag{43}$$

where $\bar{u}_k^*$ is the current displacement limit, $u_{k-1}^*$ is the displacement of the previous optimization cycle, and the subscript denotes the index of the optimization cycle. With the relaxed strategy, the stability of the optimization process is improved. Moreover, the errors of the DCPINN are discussed in the section, to better quantify the errors compared to the FE solutions, the relative errors of the nodal displacements at the specified DOF and the error norm of the displacement fields are suggested:

$$\varepsilon_{\text{dof}} = \frac{|u_{\text{PINN}} - u_{\text{FE}}|}{|u_{\text{FE}}|} \times 100\% \tag{44}$$

$$\varepsilon_{\text{norm}} = \frac{\|\boldsymbol{u}_{\text{PINN}} - \boldsymbol{u}_{\text{FE}}\|}{\|\boldsymbol{u}_{\text{FE}}\|} \times 100\% \tag{45}$$

Before discussing the errors, the optimized structures need to be analyzed to show that the displacement constraint is working. An almost identical optimization model is employed, except without the displacement constraint. Its optimized structure is presented in Figure 16 (b), and the optimized structure of another model with the displacement constraint is presented in Figure 16 (a). The entire pseudo-density field distribution reveals a similar trend, but the middle X-shaped structure is no longer vertically symmetric and a hinge-like structure appears in the bottom crossbar section. As presented in Figure 17, the displacement of the specified DOF using the volume constraint is smaller than that of the common model (without the volume constraint). These features indicate that the displacement constraint has an effect.

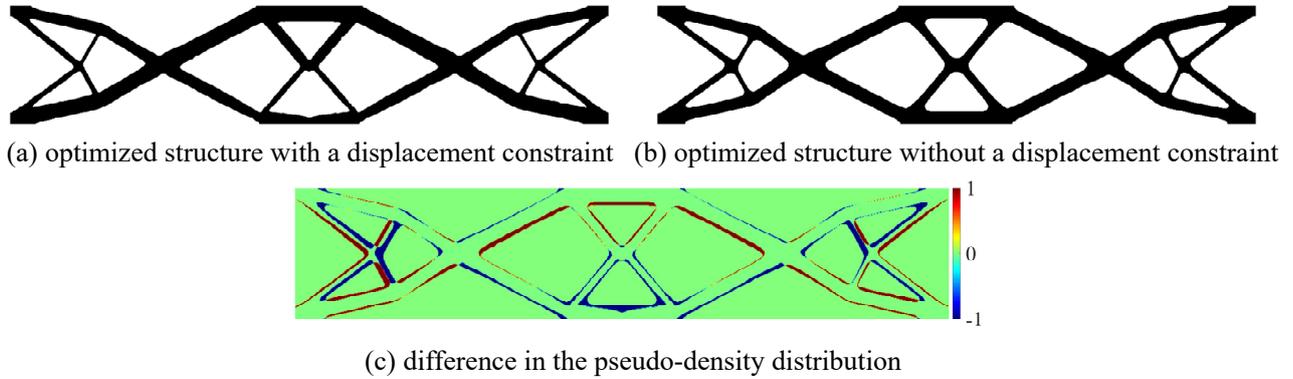

(a) optimized structure with a displacement constraint  (b) optimized structure without a displacement constraint

(c) difference in the pseudo-density distribution

Figure 16 Comparison of optimized structures, (c) is obtained by subtracting (b) from (a).

Furthermore, the accuracy of the predicted displacements is a prerequisite for the displacement constraints to work reliably. Information regarding the nodal displacement of the specified DOF is presented in Figure 17, and FE solutions with identical density fields are assessed. Clearly, the nodal displacement gradually decreases from close to 2 m until the displacement is satisfied, in which the relaxed strategy on the displacement constraint plays a key role. Furthermore, both in the relative error and the error norm, more pronounced oscillations are observed at the early stage. Nevertheless, the relative error of the specified DOF is within 3%, and the error norm of the whole displacement field is within 4%. In the remaining stage, both remained within approximately 1%. In addition, as mentioned above, the interpolation field obtained from the coefficient NN will interpolate the output of the backbone NN. As the optimization proceeds, changes in the pseudo-density field accumulate and the backbone NN is progressively less adaptable to the current pseudo-density field. Therefore, it is necessary to train the backbone NN periodically, to adapt to the pseudo-density field and avoid the accumulation of errors. The accuracy of the DCPINN models is verified throughout the optimization process, neither at the nodal displacement nor at the whole displacement field. The former explains the reliability of the displacement constraint, and the latter validates the necessity for dynamically configuring trainable parameters

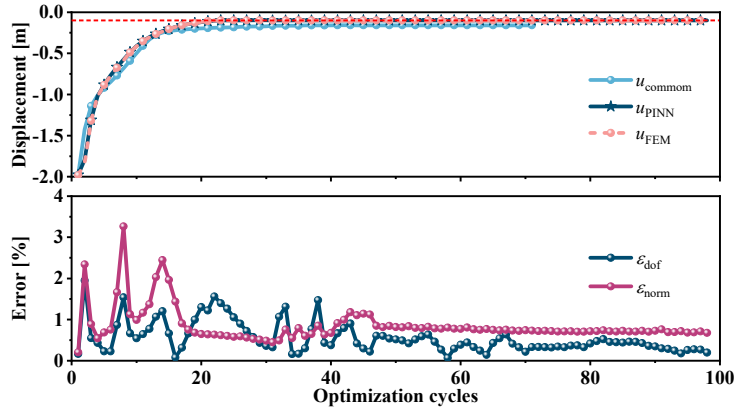

Figure 17 Relative displacement error in the constrained DOF, the first row is the nodal displacement of specified DOF, and the second row is the relative error.

Additionally, to visualize the differences in the displacement fields, the results obtained from the DCPINN models and FE methods are presented in Figure 18, along with the error distribution. Although the error occurs over a wide area, the magnitude is usually small. In fact, the maximum error is located at the boundaries of the pseudo-density field, where the pseudo-density is in an intermediate state. This phenomenon also confirms the above observation that training a DCPINN model focuses more on the collocations associated with high pseudo-density elements. Fortunately, the slight error did not cause the optimization process to break down and an elegant result is still obtained.

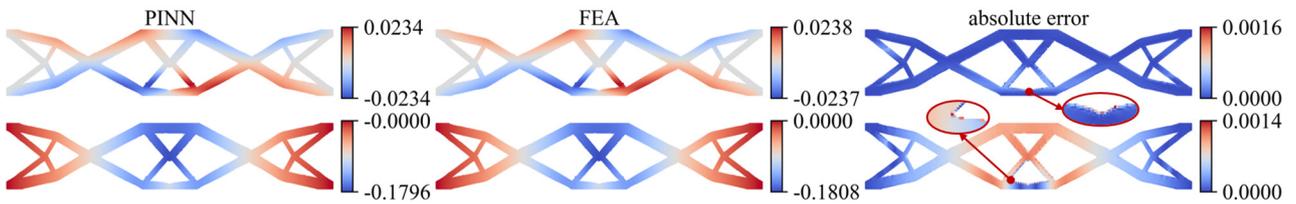

Figure 18 Absolute error distribution for displacements. The first row is associated with the X-direction, and the second row is associated with the Y-direction.

The accuracy of the DCPINN models proposed in this study is discussed, and iterative information and time consumption are discussed next. A distinction from the previous examples is that there is a large gap between the volume fraction and the predefined volume limit, which only appears early in the optimization (as presented in Figure 19 (a)). Competition between multiple constraints emerges, so that the optimization process may focus on different constraints for each iteration. As the nodal displacement of the specified DOF decreases, the optimization process adds attention to the volume constraint, so the volume begins to approach the predefined limit. The relaxed strategy acting on the displacement constraints is also, in essence, designed to prevent the displacement constraint from generating strong competing capabilities, leading the optimization to focus too much on displacement constraints. This can potentially cause the optimization to break down. Even though each optimization cycle is limited to 90% of the displacement of the previous cycle, the displacement quickly decreases and approaches the predefined limit. Eventually, it enters a fine-tuning stage, and then converges and stops optimizing. The final volume is 0.3, and the nodal displacement is $9.996 \times 10^{-1}$ m, both satisfying the predefined limits. As expected, the iterative curves of training costs also maintain the same pattern as those in the above examples.

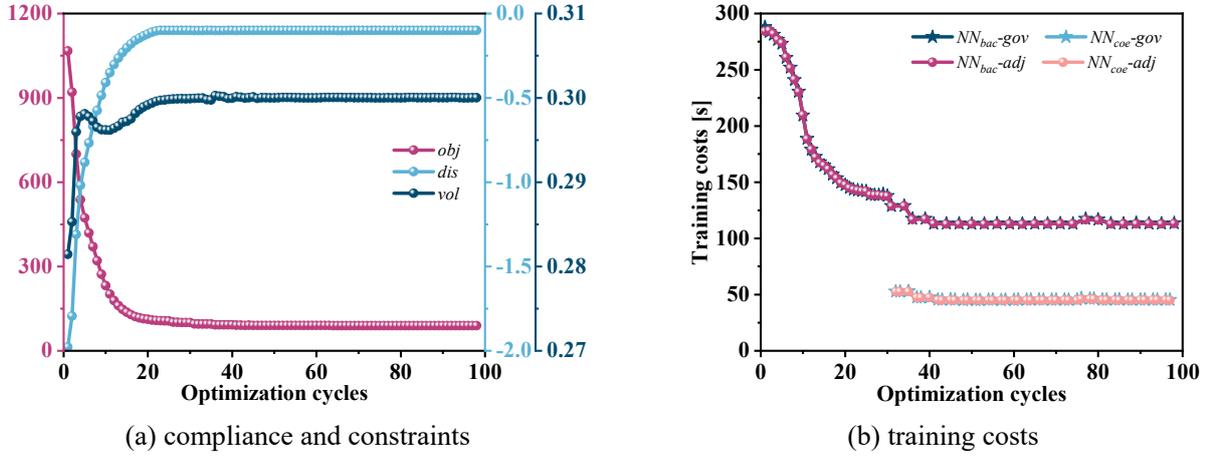

(a) compliance and constraints  (b) training costs

Figure 19 The iterative curves for the multi-constraint example.

### 3.3 Compliance minimization problem in 3D

It is well known that DCPINN models are more difficult to train when applied to 3D scenarios. In this section, a 3D cantilever beam with the size of $12 \times 1 \times 5$ m³ is discussed. In Figure 20, the left side of the structure is fixed and the top edge of the right side is subjected to a uniform force; i.e. a total of 1 kN. A circular hole with a diameter of 2.5 m was dug in the center of the structure. The design domain is discretized to a $192 \times 16 \times 80$ mesh. The volume fractions are 0.3, and the filter radius is $\sqrt{6}$ times the element size. A maximum optimization loop of 100 is set to ensure a normal exit mechanism for the routine, since the stopping criterion of $10^{-4}$ may be difficult to achieve in 3D problems. The remaining parameters are consistent with those in Section 3.1.

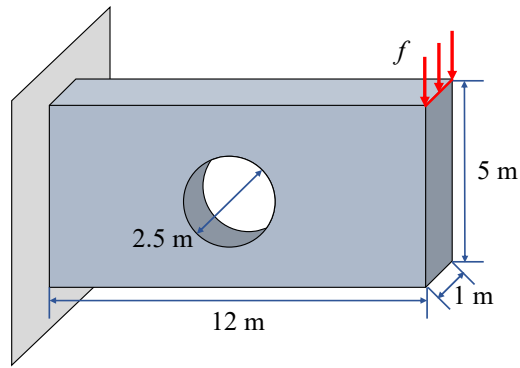

Figure 20 Design domains and boundary conditions for the 3D example.

The prediction accuracy of DCPINN is discussed in the above section, and in this subsection, the FEM-based topology optimization is introduced for comparison. The DCPINN-TO method proposed in this paper works well when addressing 3D compliance minimization problems, Figure 21 (a) illustrates the stable optimization process with smooth changes in compliance, and even the compliance of DCPINN-TO almost overlaps with that of the FEM-based approach. The relative error in compliance is calculated to quantify the difference between DCPINN-TO and FE-based TO, with a maximum relative error of only 1.94% throughout the optimization process. Notably, the active sampling strategy is effective and efficient in 3D compliance minimization problems. An interesting phenomenon is that when the optimization is executed up to approximately 40 cycles, the error in compliance decreases linearly, although this error only characterizes the optimization objective. Coincidentally, the structure stabilizes as the optimization approaches 40 cycles, which may be a function of the operation inspired by transfer learning, i.e., applying the previously trained model as the current pre-trained model. As presented in Figure 21 (b), the training costs gradually decrease to 50% of that at the high-grayscale stage. At the low-grayscale stage, the training costs of the coefficient NN are reduced by approximately 62% compared to those of the backbone NN. Compared to the 2D problems with the same number of DOFs, for more expensive 3D problems, it is clear that the two strategies can save more cost. Another advantage is that the framework of DCPINN-TO transfers from 2D to 3D problems without making complex modifications to the network architecture, simply by adjusting the dimensions of the inputs and outputs. The efficiency and flexibility demonstrate the potential of DCPINN-TO.

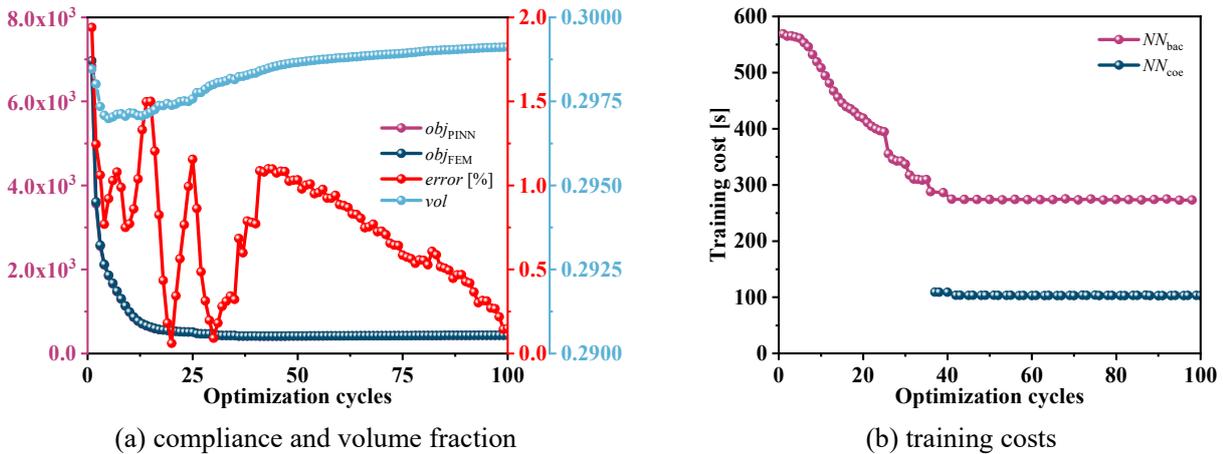

(a) compliance and volume fraction  (b) training costs

Figure 21 Iterative curves for the 3D example.

In many topology optimizations, structures with large differences in characteristics may exhibit close compliance. The structures during optimization are also presented in Figure 22 to ensure the reliability of the conclusions regarding DCPINN-TO. In addition to the final optimized structures, the structures of the intermediate iterations are also compared. For both the intermediate cycles and the final cycle, the structures of FEM-based TO are quite close to those of DCPINN-TO. Therefore, the accuracy of the predicted physical quantities of the DCPINN is considered acceptable, and at least the impact on the optimization process is weak, even for 3D problems. The optimized results demonstrate a slight increase in compliance relative to the intermediate results, but this is not caused by DCPINN, as is observed using the FEM-based approach. It may be that applying a more appropriate stopping criterion can avoid this, but this is beyond the scope of this paper.

**FEM-based**

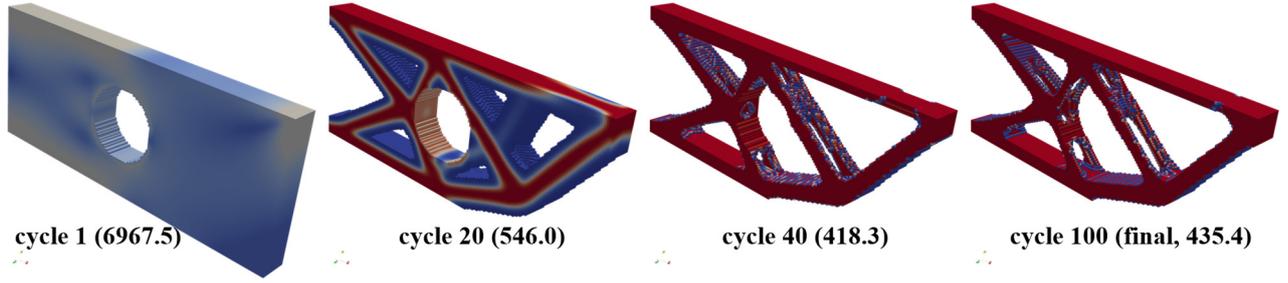

cycle 1 (6967.5)　　cycle 20 (546.0)　　cycle 40 (418.3)　　cycle 100 (final, 435.4)

**PINN-based**

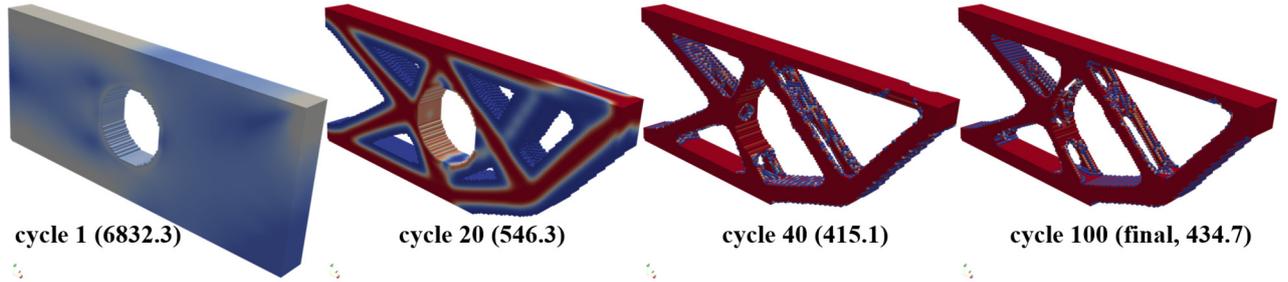

cycle 1 (6832.3)　　cycle 20 (546.3)　　cycle 40 (415.1)　　cycle 100 (final, 434.7)

Figure 22 Structures during optimization for the 3D example, the compliance in parentheses.

## 4. Conclusion

To explore the integration of topology optimization and machine learning without prohibitive data acquisition, in this paper, the dynamically configured physics-informed neural network-based topology optimization (DCPINN-TO) method is proposed. To begin with, the architecture of the CDPINN model is composed of a backbone NN and a coefficient NN, which aims to dynamically configure trainable parameters to reduce training costs. The essence of this strategy is that it leverages the similarity of the pseudo-density field in successive optimization cycles to periodically replace high-cost NNs. In addition, the active sampling strategy is proposed to select collocations according to the pseudo-density, which removes the low-contributing collocations and in turn saves training costs by reducing the amount of inputs. Moreover, critical to PINN-TO is the application of Gaussian integrals to calculate the strain energy of elements rather than that of collocations. Accurate assignment of material properties is achieved due to decoupling the displacement prediction and Young's modulus at collocations. Eventually, several examples are applied to illustrate the competence of the PINN-TO method.

To adequately demonstrate the scalability of the PINN-TO, four representative examples are employed: (**a**) common TO models with different geometries and resolutions; (**b**) a multi-load TO model with different volume fractions; (**c**) a multi-constraint TO model (add a displacement constraint); and (**d**) a common 3D TO model. In all examples, the training costs are discussed and show the effect of both the dynamically configured trainable parameters and the active sampling strategy. The savings in training costs of the active sampling strategy depend on predefined volume fractions, which are investigated with a focus on the multi-load example. However, the savings in training costs by dynamically configured trainable parameters depend on the architectures of both NNs because the number of collocations is the same. Additionally, the effect of saving training costs not only appears in single-load cases but is consistent for multi-load examples as well as the multi-constraint example. Apart from the effect of training costs, the accuracy of PINN prediction is investigated in the multi-constraint example, because of its higher requirement for displacement. The relative error of the specified DOF is within 3%, and the error norm of the whole displacement field is within 4%. In the remaining optimization cycles, both were maintained within approximately 1%. Nevertheless, the 3D example is also used to demonstrate the scalability of PINN-TO, where the maximum error in the optimization objective is only 1.94% compared to FE-based TO. In summary, the PINN-TO method proposed in this paper can solve TOs efficiently and with high accuracy and has been validated on multiple classes of problems. The above characteristics demonstrate the potential of PINN-TO proposed in this paper.

## Replication of results

The optimization problem is solved using the in-house Python implementation. The OC method is used to address the single constraint problem, and the MMA method from Prof. Krister Svanberg is used to address the multi-constraint problem.

## Declaration of competing interest

The authors declare that they have no known competing financial interests or personal relationships that could have appeared to influence the work reported in this paper.

## Data availability

Data will be made available on request.

## Acknowledgments

This work has been supported by the National Key Research and Development Program of China (No. 2022YFB3303402), Project of the National Natural Science Foundation of China (11972155), Peacock Program for Overseas High-Level Talents Introduction of Shenzhen City (KQTD20200820113110016) and Project supported by Provincial Natural Science Foundation of Hunan(2020JJ4945).